\documentclass[10pt]{article}
\usepackage{longtable}
\usepackage{tabu}
\usepackage[hyphens]{url}
\usepackage{scrextend}
\usepackage[hidelinks]{hyperref}
\usepackage{wrapfig}

\usepackage{graphicx}
\usepackage[ruled,linesnumbered]{algorithm2e}
\usepackage[numbers]{natbib}
\usepackage[singlelinecheck=false]{caption}
\DeclareCaptionLabelFormat{blank}{}

\usepackage{booktabs} 
\usepackage{authblk}
\usepackage{adjustbox}
\usepackage{color, colortbl}
\usepackage{subcaption} 
\usepackage{amsthm}
\usepackage{amsmath}

\usepackage{pdflscape}
\usepackage{enumitem}
\usepackage{lineno}
\usepackage{endnotes}
\usepackage{float}

\definecolor{Gray}{gray}{0.9}
\definecolor{White}{rgb}{1,1,1}
\definecolor{LightCyan}{rgb}{0.88,1,1}
\definecolor{orange}{rgb}{1,0.5,0}

\usepackage[space]{grffile}

\makeatletter
\g@addto@macro{\thm@space@setup}{\thm@headpunct{:}}
\makeatother

\begin{document}
\title{Automatic Large Scale Detection of Red Palm Weevil Infestation using Aerial and Street View Images}

\author[1]{Dima Kagan\thanks{kagandi@bgu.ac.il}}
\author[1]{Galit Fuhrmann Alpert\thanks{fuhrmann@bgu.ac.il}}
\author[1]{Michael Fire\thanks{mickyfi@bgu.ac.il}}
\affil[1]{Department of Software and Information Systems Engineering, Ben-Gurion University of the Negev, Israel}
    \maketitle

\begin{abstract}
The spread of the Red Palm Weevil has dramatically affected date growers, homeowners and governments, forcing them to deal with a constant threat to their palm trees.
Early detection of palm tree infestation has been proven to be critical in order to allow treatment that may save trees from irreversible damage, and is most commonly performed by local physical access for individual tree monitoring.
Here, we present a novel method for surveillance of Red Palm Weevil infested palm trees utilizing state-of-the-art deep learning algorithms, with aerial and street-level imagery data.
To detect infested palm trees we analyzed over 100,000 aerial and street-images, mapping the location of palm trees in urban areas.
Using this procedure, we discovered and verified infested palm trees at various locations.

We demonstrate that computer vision  provides an efficient and practical solution for
large scale monitoring of infested palm trees.
The results indicate that by the use of  publicly available online data
and without a need of specialized equipment, the proposed framework can be used to automatically map palm trees at  large scales and  detect ones that are potentially infested .
We show that the framework can be effective in both urban and open environments.
This method can revolutionize current old school practices for Red Palm Weevil management, proposing much cheaper cost-effective and efficient monitoring. 

\end{abstract}

%
%

	\providecommand{\keywords}[1]{\textbf{Keywords:} #1}


\begin{landscape}
\begin{figure}

\centering
\includegraphics[width=1\linewidth]{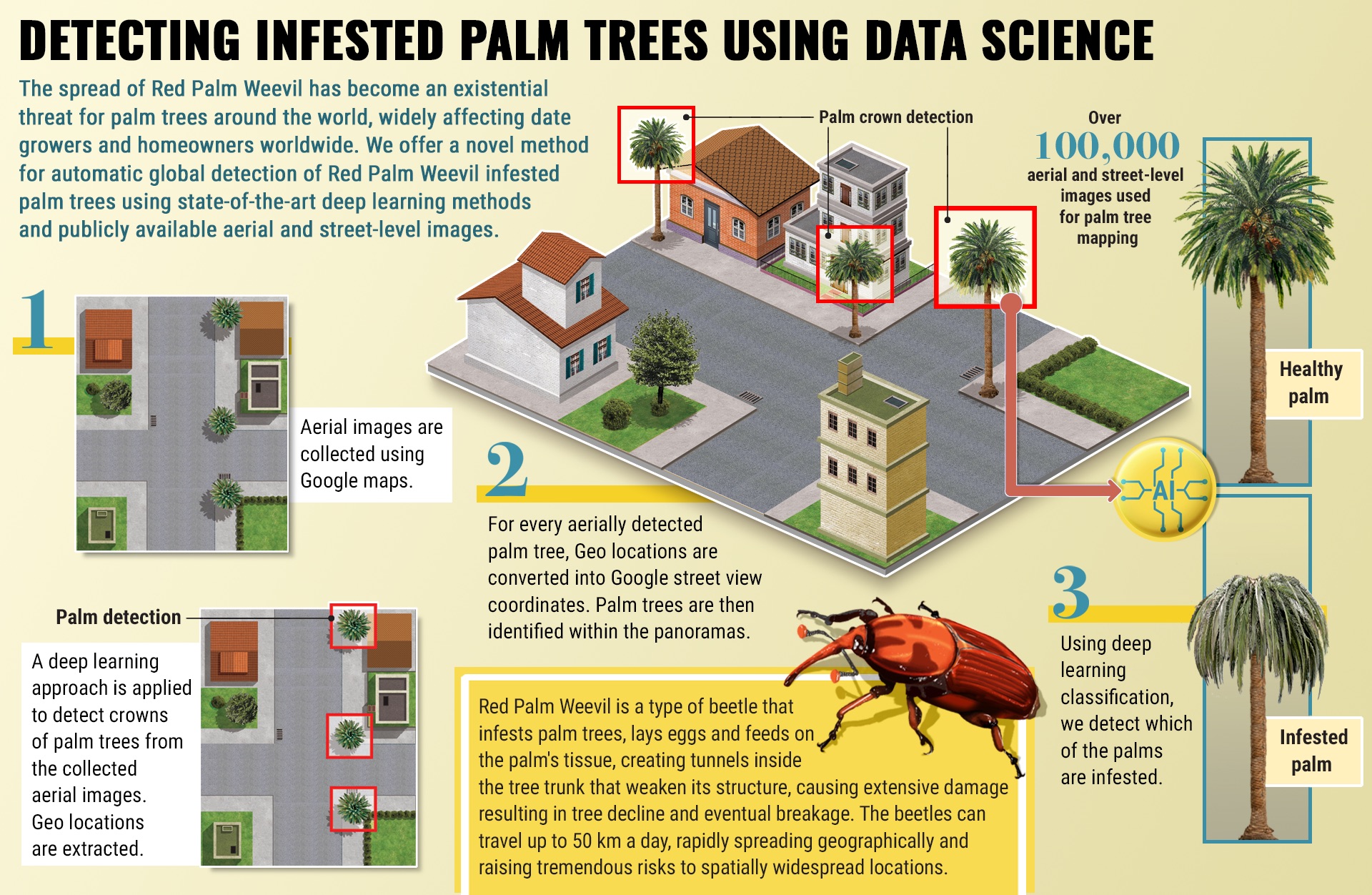}
\captionsetup{textformat=empty,labelformat=blank}
\label{fig:infog}

\end{figure}

\end{landscape}

\section{Introduction}
\label{sec:int}
\begin{figure}[t]
  \centering
    \includegraphics[width=0.8\linewidth]{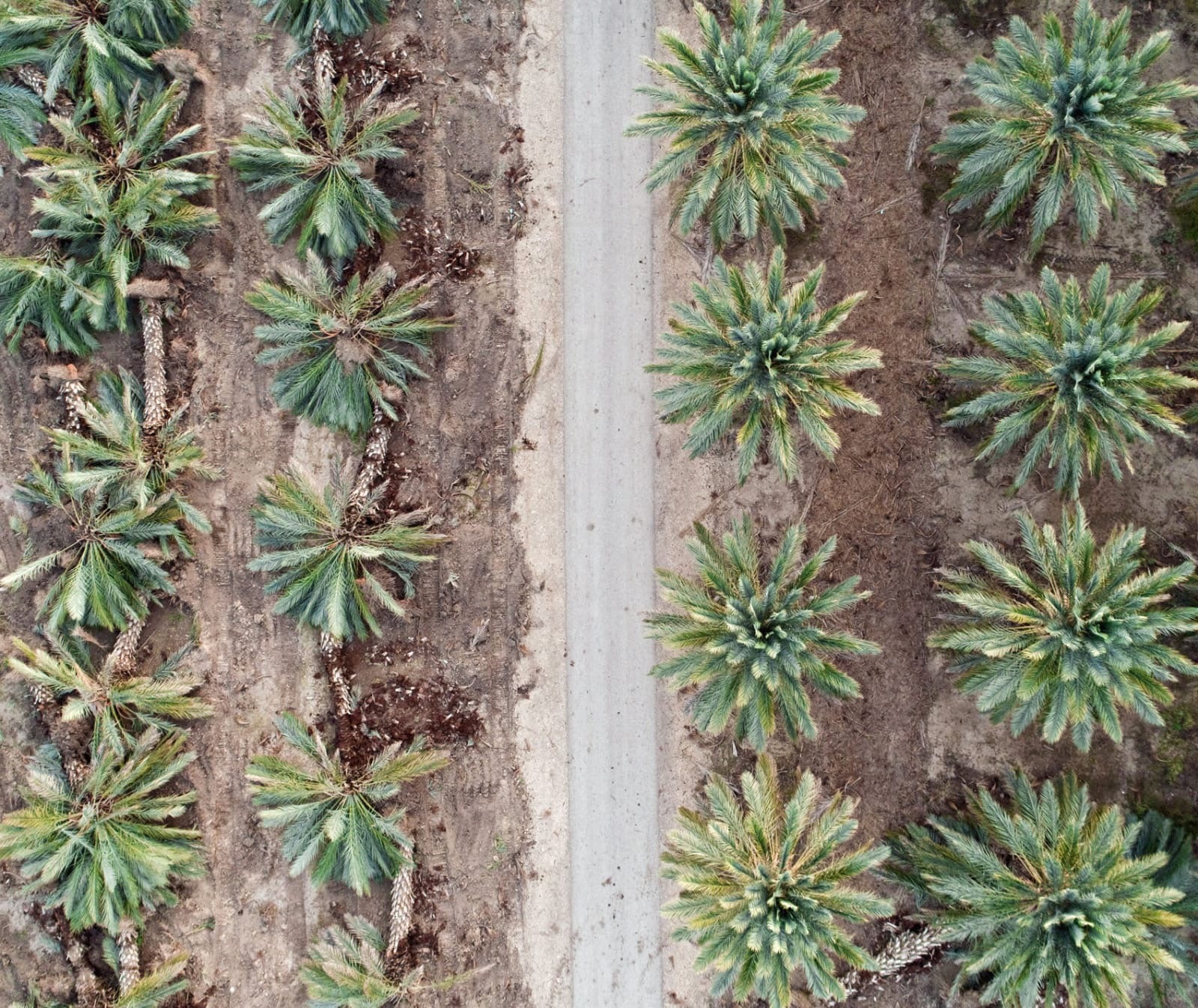}

  \caption{Deganya Alef, Israel. A Red Palm Weevil infested date plantation.
 Left side: infested trees that needed to be taken down.   Right side: trees still intact. Image  taken by Omer Tsur - AirWorks aerial photography.}
  \label{fig:infected-palms}

\end{figure}

The Red Palm Weevil (also known as Rhynchophorus Ferrugineus, and Rhynchophorus Vulneratus) is a type of beetle that attacks palm trees and that has become an existential threat for palm trees around the world.
The mechanism of infection involves the beetles laying eggs inside palm trees, with the eventual larvae feeding on the palm's tissue, thus creating tunnels inside the tree trunk that weaken its structure, finally causing extensive damage that results in decline and even breakage of the tree (see Figure \ref{fig:infected-palms}).

Importantly, the spread of beetles from one palm to another is estimated up to 50 km per day~\citep{hoddle2015far}, thus rapidly spreading geographically and raising tremendous risks to spatially widespread tree locations.
In fact, the Red Palm Weevil has originated from tropical Asia areas \citep{Rhynchop57:online}, yet in the past decades, has evolved from being only a local problem into a worldwide concern.
In 2010, the Red Palm Weevil reached the U.S \citep{LatestPe76:online}, which in the following year was also invaded by its close relative species- the South American Palm Weevil \citep{USDADete29:online}.
In 2011, it was already detected in eight European countries, and it has spread further in the past decade \cite{eu:online}.
Today, according to the EPPO (European and Mediterranean Plant Protection Organization) data sheet, the Red Palm Weevil has spread to 85 different countries and regions worldwide \citep{EPPO}.
Without constant monitoring, the Red Palm Weevil will keep spreading even further.

The financial implications of the extensive damage to palm trees are enormous, particularly to date and coconut growers.
In fact, the Food and Agriculture Organization of the United Nations estimates that in 2023, the combined cost of pest management and replacement of damaged palm trees, in Spain and Italy alone,  will reach 200 million Euro \citep{RedPalmW92:online}.
Moreover, the threat from the Red Palm Weevil occurs to be not just financial, but also of injury to matter and individuals.
In many countries around the world, palm trees are considered decorative trees planted in residential neighborhoods, and an infested tree has a higher likelihood of breaking down under strong winds, thus endangering human lives.
For example, in Israel, the  Ministry of Agriculture has officially stated that it is just a matter of time until someone would be seriously injured from a collapsing infected palm \citep{TheMinis3:online}.

Importantly, early detection of palm tree infestation may save trees from irreversible damage \citep{2020FAO}.
At those earlier stages, control methods could be used, most commonly by the application of insecticides \citep{2020FAO}.
Thus, to eradicate the pests efficiently, an accurate mapping of infestation hot-zones is highly advisable, especially of privately planted palm trees, whose locations are not documented officially.
In the past couple of decades, various Red Palm Weevil detection methods were proposed.

There have been large efforts based on acoustic detection \citep{siriwardena2010portable, rach2013design, mankin2011recent, mankin2016acoustic, pinhas2008automatic, ashry2020early} as well as methods  utilizing canines \citep{nakash2000suggestion, suma2014use}.
However, these methods do not deal with the mapping of tree locations.
Moreover, most of these attempts fall from being applicable
at a large scale as well as in urban areas, where many homeowners are unaware of the danger and do not realize infestation until it is too late for treatment \citep{RedPalmW37:online}.

In this paper, we present a novel method for large scale mapping and detection of Red Palm Weevil infested palms  
using state-of-the-art deep learning algorithms.
We believe our proposed large scale approach may be of highly financial importance to countries around the world, reduce risks of injury at urban areas, and massively save agriculture fields.

Our algorithm is based on the following steps:
First, we collect and label aerial and street-level images of palm trees.
We then train three deep-learning models for palm detection on aerial images, palm detection from street-level images, and finally an infested palm classifier on the detected images.
Next, we utilize the trained aerial palm detection model to identify palm trees downloaded from Google Maps aerial data and map the identified palm tree locations into spatial coordinates.
For each detected tree we retrieve the nearest street view panorama from Google Street View and compute the camera heading that centers on the palm tree detected in the aerial images.
These tree centered images are used for the detection of palms by our street-level palm detection model.
Finally, we apply a novel infested palm tree classifier
to uncover infested Red Palm Weevil palm trees and highlight hotspots of palm trees, with a high probability of tree infestation.

We analyzed more than 100,000 images mapping palm trees utilizing both aerial and street-level imagery. We demonstrate that a combination of aerial and street-level imagery produces a cost efficient method for mapping specific objects. We apply deep-learning based image processing algorithms to demonstrate that it is possible to identify infested palm trees based on low quality street view images.
Additionally, we show that in many cases it possible to monitor the progress of the infestation. We demonstrate that there is an opportunity to revolutionize Red Palm Weevil management using computer vision.

The multiple key contributions of this study are the following:
\begin{itemize}
    \item A novel framework (see Section~\ref{sec:method}) that utilizes deep learning and imagery data to automatically detect Red Palm Weevil infested palm trees quickly over large geographic areas.
    
    \item An approach for palm tree mapping in urban areas using aerial or street-level images that allows municipalities to efficiently (quickly at low cost) perform preventive chemical treatments.
    We demonstrate that we can monitor palm tree degradation and inspect changes in the status of infestation over time. 

    \item The presented approach can be utilized to 
    identify regions in which there is a higher probability for infestation, and as a result, should be surveilled more carefully. 



    


\end{itemize}

The remainder of the paper is organized as follows: In Section~\ref{sec:rw}, we present an overview of related studies. In Section~\ref{sec:method}, we describe the datasets, methods, algorithms, and experiments used throughout this study. In Section~\ref{sec:results}, we present our results. In Section~\ref{sec:dis}, we discuss the obtained results. In Section~\ref{sec:lim}, we present the research limitations. Lastly, in Section~\ref{sec:con}, we present our conclusions from this study.

\section{Related Work}
\label{sec:rw}

In this study, we offer a novel framework to automatically detect Red Palm Weevil infested palm trees on large scale geographic areas, that is also applicable for urban regions. Both these points are currently not supported using existing methodologies.
We therefore cover here work related to both these novel points offered in the current manuscript, namely- large scale detection and urban mapping.

\subsection{Large Scale Detection of Red Palm Weevil Infestation}
As noted in the Introduction, early detection of palm tree infestation is critical in order to allow treatment that may save trees from irreversible damage.
There are various methods for Red Palm Weevil infestation detection, 
all of which are geographically limited by scale \citep{soroker2013early}.
The most straightforward method is by visually inspecting tree by tree.
However, this type of method is naturally not feasible on a large scale \citep{soroker2013early}.
To overcome this limitation, other methods were proposed.
One such possibility that was raised was of using trained animals for odor-based detection.
Specifically, research indicated that insects emit chemicals into the air \citep{allen19992011} and that dogs could be successfully used for detecting Weevil infested palm trees \citep{nakash2000suggestion, suma2014use}.
However, this approach too, could not be applicable at large scales \citet{soroker2013early}.
Another approach that was raised is sound-based detection, relying on the fact that the Red Palm Weevil larvae produce sounds while feeding through the palm tree.
Multiple such frameworks were suggested over the years \citep{siriwardena2010portable, rach2013design, mankin2011recent, mankin2016acoustic, pinhas2008automatic, ashry2020early}.
There is no doubt that the acoustic method is feasible, yet it too requires approaching each tree individually \citep{soroker2013early}, thus limiting its scale of performance.
There are indications that in some conditions, yet another detection approach, based on thermal imaging could be used to detect infested palm trees \citep{soroker2013early, golomb2015detection}. 
However, at its current state it is only suitable for detection of infestations in open areas and its detection accuracy is lower than detection by either dogs or acoustic based approaches \citep{ahmed2019detection}.
In summary, monitoring palms at a large scale is considered a challenging and presumably costly task, especially in urban areas where there is a large variance in age, spices, growth, and condition of vegetation \citep{soroker2013early, mohammed2020recent}.

\subsection{ Mapping Objects in Urban Environments Based on Street View Images}
Google street view has been used in multiple applications for urban street mapping, with applications to a variety of domains.
Below is a brief introduction to several such sample studies.
In 2015, \citet{balali2015detection} created an inventory of street signs using imagery data collected from Google Street View.
To detect and classify the collected dataset of signs, they used HOG (Histogram of Oriented Gradients) along with color,
using a linear (SVM) classifier.
In a follow-up study, \citet{campbell2019detecting} used deep learning for mapping street signs 
in a sample city (The City of Greater Geelong), demonstrating that 
the model can automatically learn to detect two different types of street signs (Stop and Give Way signs). 
In 2017, \citet{gebru2017using} used 50 million Google Street View images to estimate the socioeconomic status of 200 US cities.
They used deep learning to detect the make, model, and year of vehicles in each area.
The data analysis revealed an interesting correlation between car information and demographic data, including voter preferences, ethnicity, etc.
The same year, \citet{seiferling2017green} used Google street view to quantify tree coverage in urban areas.
To estimate tree coverage, they used a multi-step image segmentation model on the collected data.
Along those lines, in 2018, \citet{branson2018google} presented a method for mapping the actual tree species using street imagery.
They combined both street and aerial imagery to train a CNN (Convolutional Neural Network)-based classifier to classify tree images into 140 different tree species. 
Their important work emphasizes the vast potential of using advanced technologies on online data sources in order to map and classify vegetation in urban areas.
In another interesting study of urban mapping, published in 2019, 
\citet{helbich2019using} used Google Street View, satellite imagery, and deep learning to map green (vegetation) and blue (waterscapes) spaces in Beijing in order to explore correlations between green and blue spaces to mental health among the elder population.
\citet{law2019take} used street view and satellite imagery, and deep learning to extract visual features from satellite and street view images, in order to estimate housed prices.
Their research was driven by the assumption that house prices are affected by the neighborhood amenities and other features that can be quantified using images.

These and other studies demonstrate the tremendous potential of urban mapping for an amazingly wide range of applications, including those of tree mapping, as the one offered in our study.

\section{Methods and Experiments}
\label{sec:method}

\subsection{The Proposed Method and Training of the Models}
\begin{wrapfigure}{r}{0.45\textwidth}
  \begin{center}
    \includegraphics[width=0.43\textwidth]{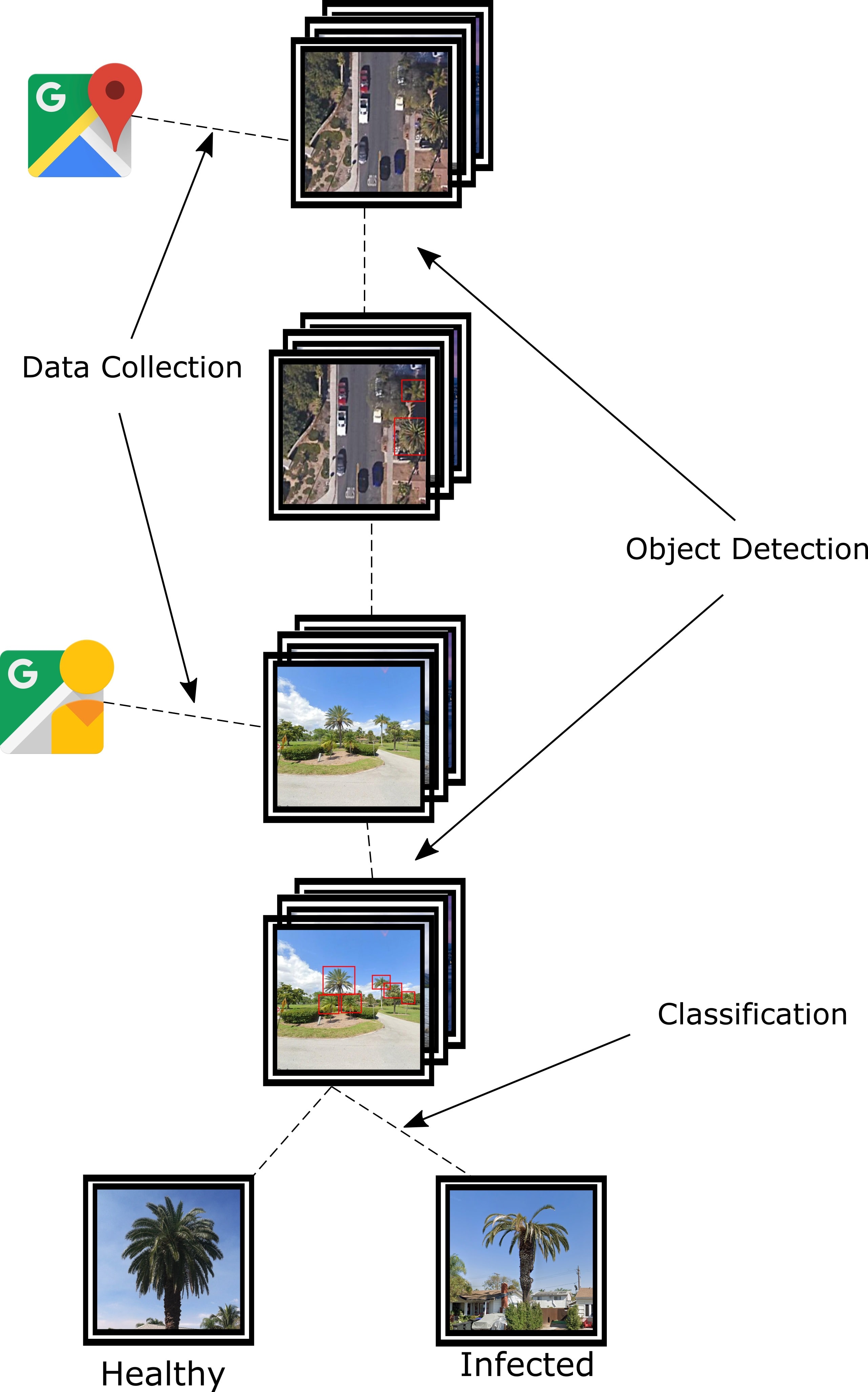}
  \end{center}
  \caption{Infested palms detection process.}
  \label{fig:flow}

\end{wrapfigure}
In this study, we strive to demonstrate the potential of automatically detecting Red Palm Weevil and South American Palm Weevil infested palm trees at large scales using computer vision algorithms.
The rationale is to provide a basic 2-step approach in order to map all palm trees in a given area, using a tree detection model and a subsequent classifier to identify infested trees amongst them.
Such an approach would be most suitable for small cities or cities highly populated with palm trees.
However, it would be inefficient for large cities, or cities with relatively low densities of palms.
In fact, since collecting and processing images may carry considerable costs in terms of both money and time, the efficiency of the palm-tree detection method in use is utterly important.

Thus, to improve  efficiency of  the proposed method, we added an additional pre-processing step prior to 
palm tree detection from street-level images.
In this pre-processing step, we first detect palm trees from aerial images and only then proceed to detection from street level,  the advantage being that while street-level imagery can be used for object detection in a radius of several meters, a single aerial image can be used for preliminary detection of objects in much broader areas.
This allows an efficient large scale scanning of areas, taking advantage of the trees being large enough objects to be detected from satellite images \citep{branson2018google}.
Importantly, aerial detection provides precise coordinates that can later be used for flagging towards treatment actions. This is in contrast to Street View imagery, which provides coordinates of the camera in use, along with
the camera heading, but not the actual coordinates of the detected object.

\begin{figure}
\captionsetup[subfigure]{justification=centering}

    \centering
    \begin{subfigure}[t]{0.3\textwidth}
        \includegraphics[width=\textwidth]{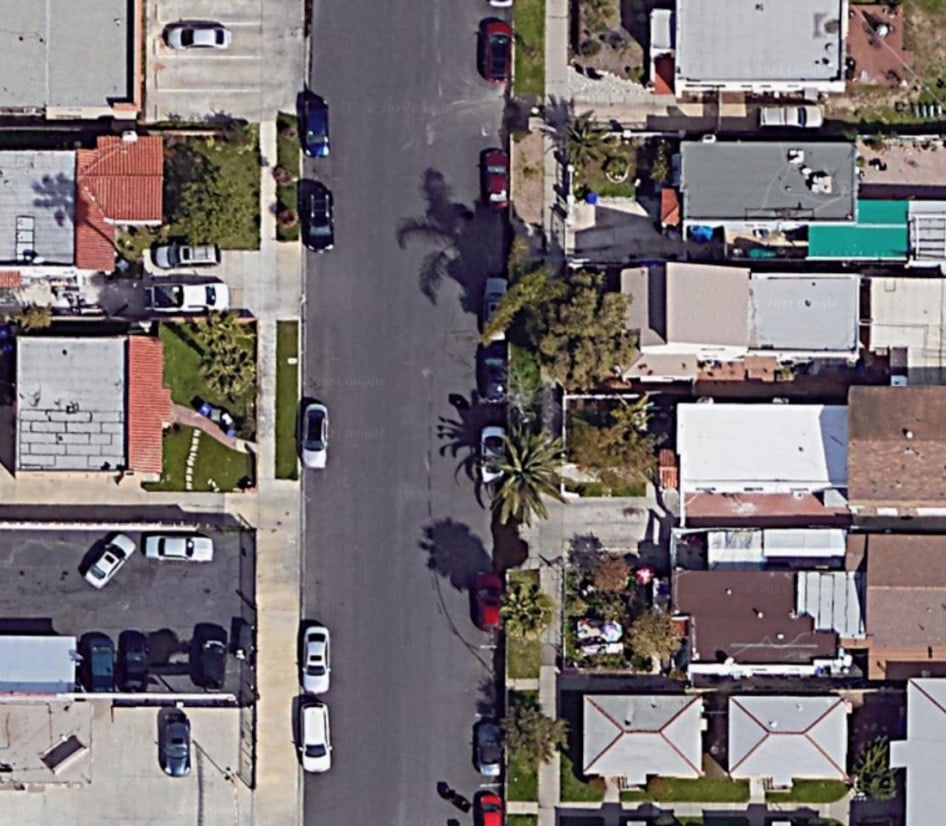}
        \caption{An aerial image.}
        \label{fig:aerial}
    \end{subfigure}
~
    \begin{subfigure}[t]{0.3\textwidth}
        \includegraphics[width=\textwidth]{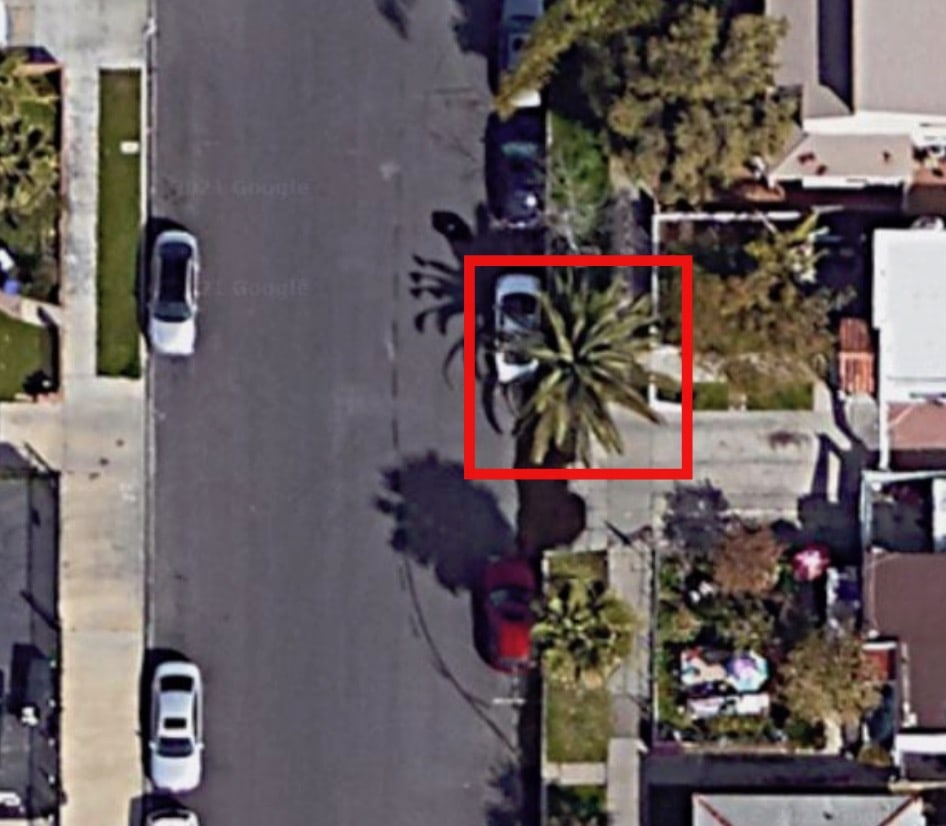}
        \caption{Palm detection in an aerial image.}
        \label{fig:aerial-detection}
    \end{subfigure}
~
    \begin{subfigure}[t]{0.3\textwidth}
        \includegraphics[width=\textwidth]{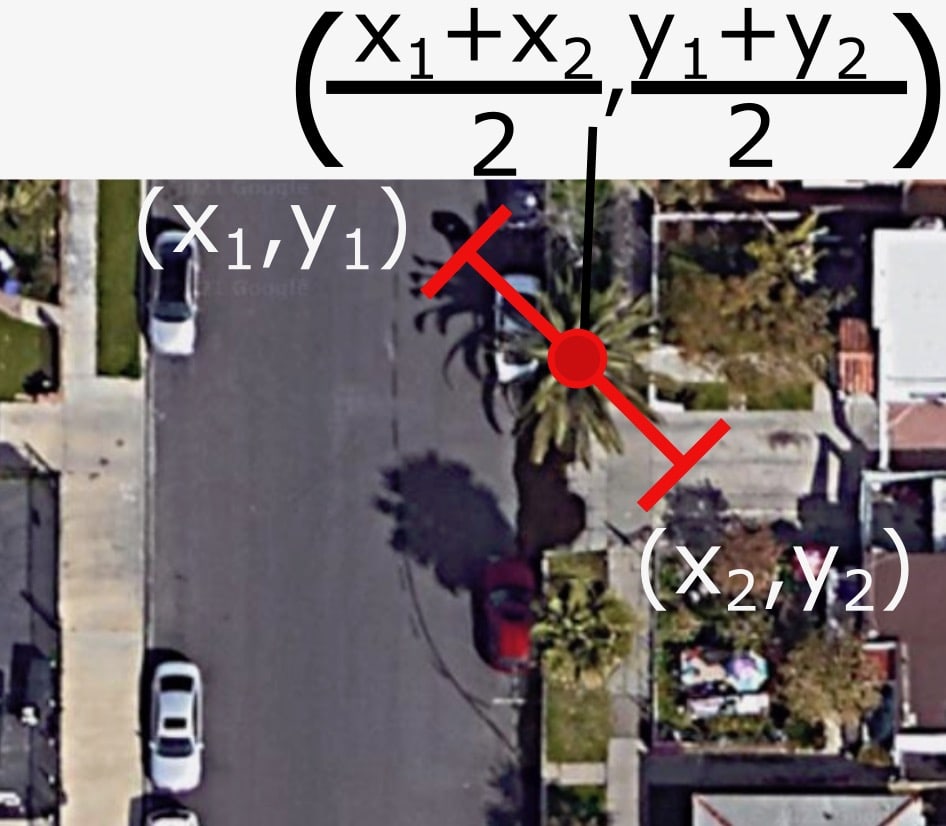}
        \caption{Calculating palm coordinates.}
        \label{fig:middle}
    \end{subfigure}
~
        \begin{subfigure}[t]{0.3\textwidth}
        \includegraphics[width=\textwidth]{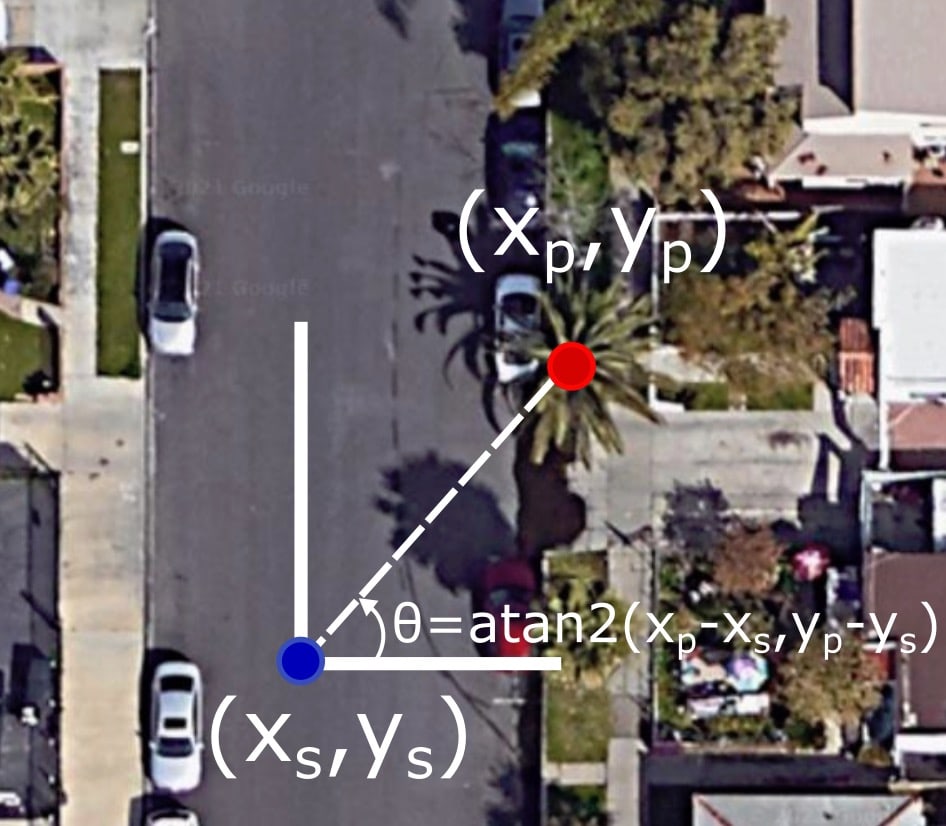}
        \caption{Calculating  angles between the streetview image and the palm.}
        \label{fig:angle}
    \end{subfigure}
~
        \begin{subfigure}[t]{0.3\textwidth}
        \includegraphics[width=\textwidth]{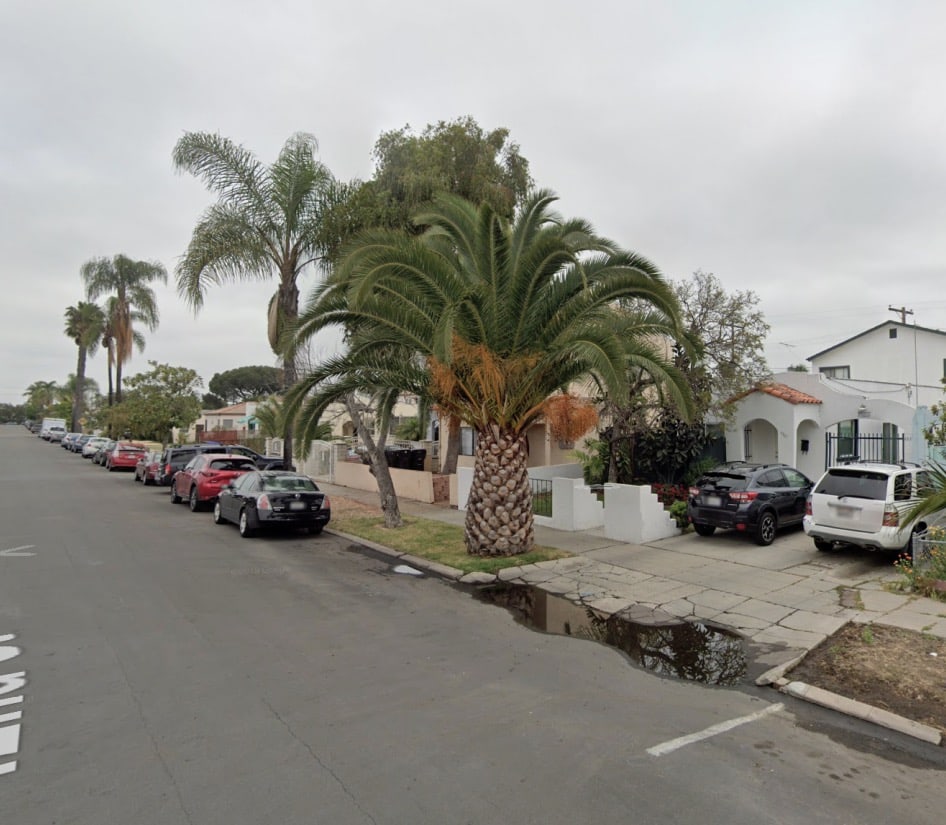}
        \caption{Retrieving the  street view image nearest to the palm tree.}
        \label{fig:street}
    \end{subfigure}
~
    \begin{subfigure}[t]{0.3\textwidth}
        \includegraphics[width=\textwidth]{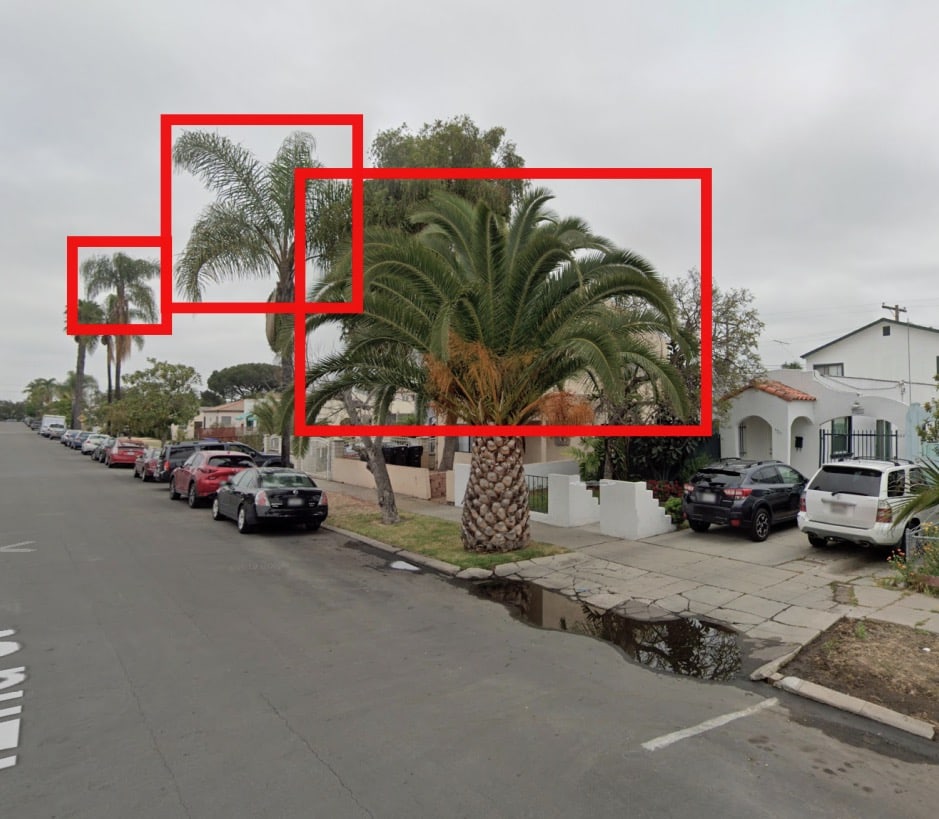}
        \caption{Palm tree crown detection.}
        \label{fig:mouse4}
    \end{subfigure}
    \caption{Palm tree detection process step by step.}\label{fig:stree-detection}
\end{figure}

The proposed method is thus composed of the following steps:
\begin{enumerate}

\item \textbf{Palm Tree Detection from Aerial Imagery}

\begin{itemize}

\item \textbf{Data Collection.} To train a palm tree detection model we collected a set of 257 aerial images containing palm trees. 
We focused on images collected from Miami area in FL (US), an area known to have high densities \citep{2tax22Pa95:online} and rich varieties of palm trees \citep{Ourmight55:online}.
The images were collected from the Miami Dade county imagery.

\item \textbf{Creating a Training Set.} We created a training set by manually labeling 257 images containing a total of 1,028 palm trees. 

\item \textbf{Training an Aerial Palm Tree Detection Model.} We used the PyTorch\footnote{\url{https://pytorch.org/}} library to apply transfer learning on a pre-trained Convolutional Neural Network (CNN). 
In order to achieve a better generalization of the model in use, we increased the variance of the data by data augmentations in multiple dimensions.\footnote{The following parameters were augmented randomly: hue, saturation, brightness, contrast, RGB shifting, vertical flip, horizontal flip, gaussian blur, rotation.}
The model was pre-trained on the COCO dataset, the standard training dataset used by PyTorch for object detection and segmentation problems. 
From the object detection pre-trained models provided by PyTorch, we chose to use Faster R-CNN ResNet-50 FPN \citep{lin2017feature}, which is an improved version of the model that achieves higher Average Recall (AR) and Average Precision (AP) without sacrificing speed, or memory and offers the fastest performance in terms of training and inference time \citep{torchvis13:online}.
Training and inference time is critical since we use a single RTX 2080 GPU on over 100,000 images.
The model was evaluated using the Mean Average Precision\footnote{Pascal VOC metric} (mAP) metric on a  20\% validation set.

\item \textbf{Extracting Coordinates of Tree Object Locations.}  Using the trained aerial detection model, we detected trees over wide urban areas (see Figure \ref{fig:aerial-detection}).
To convert the output of the detection model from (x,y) coordinates, corresponding to  image pixels, into actual physical locations, we performed the following:
First, since input images are given in the format of tiles, and tile border coordinates per image are known, we mapped tile coordinates from Google coordinate format (EPSG:3857) onto WSG 84 coordinate system and calculated the bounds of the tile.
We then performed affine transformation to convert the detected palm trees bounding box boundaries onto WGS 84 coordinates.
Finally, we calculated the center of the bounding box to represent the coordinates of a palm tree location (see Figure \ref{fig:middle}).

\end{itemize}

\item  \textbf{Palm Tree Detection from Street View Images in Urban Environments:}

We trained an object detection model to detect palm trees from the collected street-level images.

\begin{itemize}

\item \textbf{Data Collection.} 
To train a street view palm detection model we first 
collected a set of 314 Street View images of palm trees.
Images were collected using Google Street View, each image size was 640x640.
In accordance with the collected data for the aerial detection model, we focused on images collected from the same region of Miami-Dada county, FL.

\item \textbf{Training the Street View Palm Detection Model.} 
We performed manual labeling of palms on the collected dataset, by drawing bounding boxes over palm tree crowns while minimizing background as much as possible.
A total of 314 images, containing 888 palm trees, was used for training the model.
Similarly to the aerial detection model, also here, we applied transfer learning using Faster R-CNN ResNet-50 FPN \citep{lin2017feature} pre-trained on COCO as our model,
with the PyTorch\footnote{\url{https://pytorch.org/}} library.
Data augmentations were also performed as for training the aerial detection model.
The model was evaluated using the mAP metric on a 20\% validation set.

\item \textbf{Localizing Palm Trees for Classification.} 
To determine the physical location of the palm tree for subsequent classification, for each palm tree detected in the aerial images, we extracted the nearest coordinates of a street view panorama.
Next, we calculate the required heading of the camera using the following equation (see Figure \ref{fig:angle}): 
$$fov=atan2(x_{pano}-x_{aerial},y_{pano}-y_{aerial})$$
Using the calculated $fov$ for each palm tree we retrieved the corresponding street view image (see Figure \ref{fig:street}).
We then applied the palm tree crown detector to identify all the palm trees in the image (see Figure \ref{fig:stree-detection}).

\end{itemize}

For each palm tree crown detected by the trained street-level palm crown detector, we proceed to predict whether it is healthy or infested.

\item  \textbf{Classification of Detected Palm Trees into Healthy/Infested:}

To train a model for identifying whether detected palm trees are healthy or infected, we used 70 images containing infested palm tree crowns.
We used only the palm crowns for classification, the logic being that since the resolution of palm crowns is relatively low in Street View images (less than 640x640\footnote{The max size of a Street View image}) it should be easier to detect infestation symptoms in crowns than in the tree trunk in low resolution images.
We used the same images that were manually labeled for training the palm crown detection model. 
During the process of data, we realized that the Palm Beach data did not have samples of Date Palms.
Thus, to add Date Palms to the data, we labeled additional 224 healthy palm trees from the rich in palms neighborhood in Omer, Israel, and 53 healthy palm trees from Los Angels, US.

We used transfer learning on the XResNet model pre-trained on ImageNet, with the fastai \citep{howard2020fastai} library.
XResNet is an improved ResNet architecture developed by fastai and based on the work of \citet{he2019bag}.
XResNet features three tweaks (ResNet-B, C, and D) which \citet{he2019bag} demonstrated to improve model accuracy consistently.
We used fastai standard augmentations and progressive resizing \citep{howard2020deep} for training the model.

To train the classifier, we integrated out-of-domain data, namely images not containing palm trees.
Since a classifier will always return  a class with the highest value as classification we included an unknown class to contain such out-of-domain data samples.
In fact, \citet{zhang2017universum} has shown that such an approach has an extra regularization effect with supervised learning.
To get a variable sample of out-domain-data we used the Caltech 101 dataset \citep{fei2006one}.
This dataset contains 101 different categories to be used for the ``unknown'' class.
We also added object data samples collected from street view images, including non-palm trees, billboards, cars, etc.
To deal with data imbalance, the infested palm trees were over-sampled (from 70 to 892) and the Caltech 101 classes were under-sampled uniformly 8 images of each class. 
We tested the model on palm trees that we manually identified as infested and on images of the same palms prior to infestation.
To manually identify infested palm trees we relied on the Food and Agriculture Organization (FAO) guidelines \citep{2020FAO}.
According to the guidelines, infested palm trees should have symptoms such as oozing of a
brown, viscous liquid from the site of infestation, boreholes, large cavities, and even the crown can fall off.
Different palm trees have different symptoms.
Phoenix canariensis specifically and some types of date palms should have symptoms that are more visible in the crown.
For example, they would have holes in the fronds, absence of new fronds,  wilting/dying of already developed fronds, and/or an asymmetrical crown.
Also, sometimes leaves above the older leaf whorls are dry.
Limited by the image resolutions of Google Street View, we focused on detecting symptoms that are manifested in the palm tree crown.

\end{enumerate}

\subsection{Curating an Image Datasets}
\label{sec:dataset}
We curated data for three different tasks:
\begin{enumerate}
    \item Palm tree detection from aerial imagery: To collect aerial imagery data we used Google Maps. 
    The tiles (images) were collected for a specific geographic area at zoom 20. 
    The image size was 256x256 pixels.
    \item Palm tree detection from street-level: To collect street-level data we used Google Street View.
    Each image has a heading of 90 degrees.
    The image size was 640x640 pixels.
    \item Infested palm tree classification: To collect healthy and infested palm tree images we manually collected images for both classes. 
    Infested palm tree images were collected from various websites utilizing Google Search.
    Images that represent healthy palm trees were selected manually using images downloaded from Google Street View.
\end{enumerate}

\subsection{Evaluating the Method on Test Data}

For demonstrating the potential of our proposed method, we focus on two physical locations reported of having infested palm trees.
For evaluation of the method, we choose to focus on the San Diego 
area since according to a report by \citet{hodel2016south} in 2016, an infestation was found in the San Ysidro area of San Diego, and since there are also street view images from the referred period.
As another case study for palm tree mapping we chose the small town of Omer, Israel, which contains a high density of palms.
Because of its small size and its high density of palm trees, it is a perfect use-case for demonstrating tree mapping from street-level images.

To test  performance of our palm mapping methodology, we evaluated three sub scenarios- aerial mapping, street-level mapping, and combined method.

To test  performance of our infested palm trees detection classifier in urban areas, in real-world conditions, we collected street image data from the San Ysidro, Imperial Beach, Old Town, Petco Park areas.
The images for San Ysidro were collected between February 2015 and April 2016, a time period prior to the date specified in the report \citep{hodel2016south}.
For the other areas, the images were collected from 2018 according to the timestamp on the presented images in the website \citep{SouthAme35:online}.
We used the street layer supplied by an open street map to extract coordinates that represent the streets. The coordinates were extracted with a distance of eight meters between each following point.
For each point, we collected the nearest panorama from which we extracted four images in the fields of views of 0, 90, 180, and 270.
We then used our palm tree classifier to extract all the palm crowns from the collected street view images and finally classified the crowns to either healthy or infested.
To evaluate performance, we searched for the trees presented in the reports \citep{hodel2016south, SouthAme35:online} as well as
additional potentially infected trees.

To blindly find newly infested trees for which we had no prior reports (see Figure \ref{fig:flow}), we used our full method as presented in the Methods section, by initially detecting the palm trees using aerial imagery, then applying detection and classification on Street View images.
Since the search space is enormous, as a proof of concept, we focused on the San Diego area, specifically on neighborhoods where infested trees were found in the past.
The usage of aerial images reduces the search space saving time and money.
We also inspected additional neighborhoods where residents mention Red Palm Weevil on social media.
These raised only recent street view photos from 2019 and 2020 from which we retrieved the newest photos for each location.

To measure the effectiveness of using aerial imagery for reduction of the search face we performed a comparison of both methods.
We calculated the required number of street view images to fully, map a specific area.
Next, we used the full method on the same area and calculated the number of aerial and street view images the was required to map the same area.

Finally, we also explored the temporal propagation of palm tree infestation, in search of the point in time in which a palm tree was infested.
We chose the palm trees that were classified as infested and retrieved their history street view images.
Since each image may contain multiple trees and repeated street view images are not necessarily taken at the exact same coordinates, we employed the following heuristics to re-center the trees. 
First, we calculate the point of view using the newer street-view coordinates and the aerial coordinates using $atan2$.
If the newer street viewpoint is farther from the aerial retrieved coordinates than the original street view image we retrieve the image using the calculated point of view.
Otherwise, we calculate the shift of the palm in the image ($90*((x_{left}+x_{right})/2)/640 -45$) and it to the calculated heading.
Next, we classified the palm tree past images to detect at which times it was already infested.

\section{Results}
\label{sec:results}

\begin{figure}[h!]
\includegraphics[width=\textwidth]{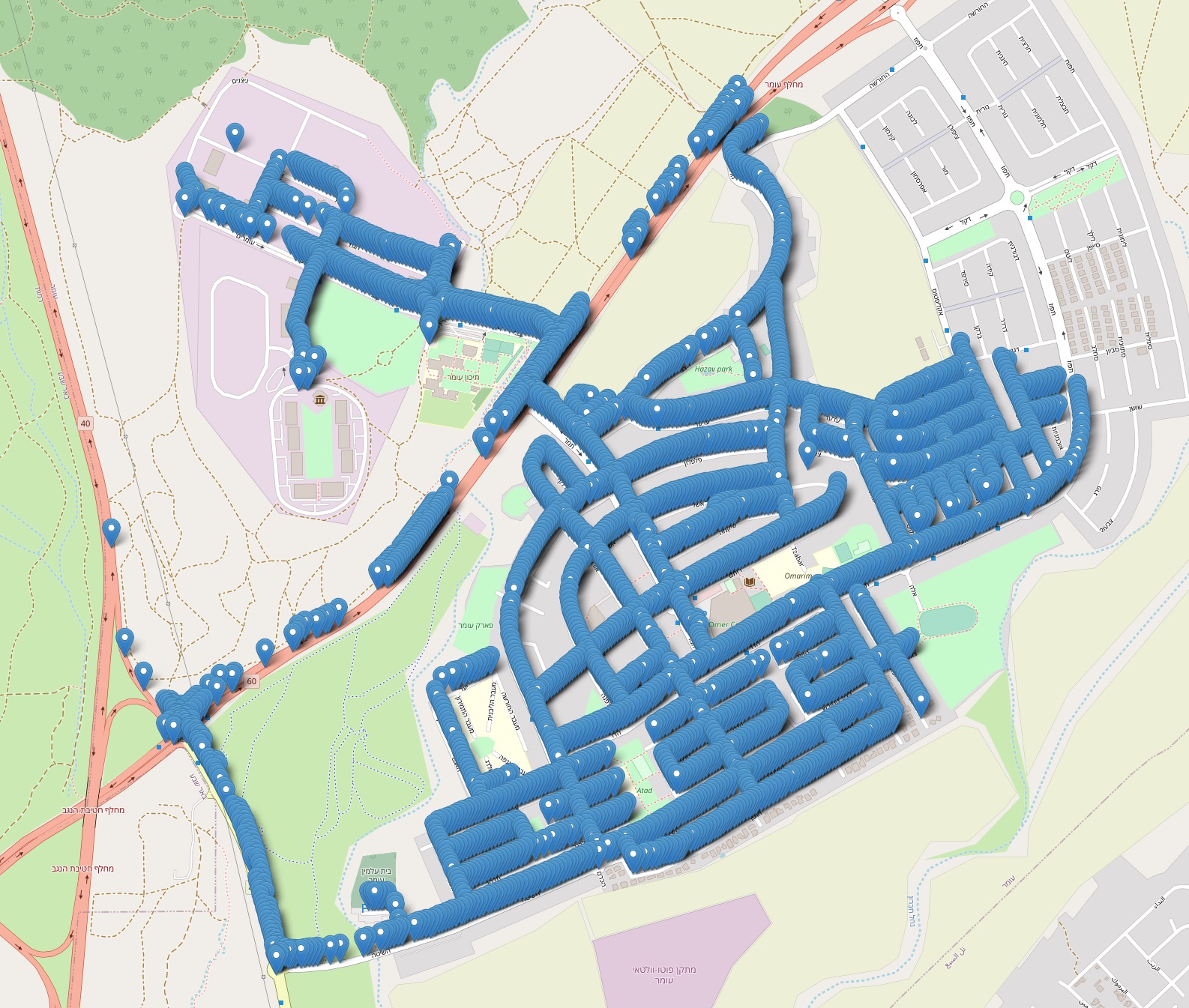}
\caption{Streetview panoramas locations in which palm trees where detected.}
\label{fig:omer}
\end{figure}
To detect and map Weevil infested palm trees (see Section~\ref{sec:method}), we collected over 100,000 aerial and Street View images that were online available on the Google Maps platform.
Out of the downloaded dataset, we extracted a total of 47,138 aerial  61,009 street-level images of palm trees. 
Using our proposed methodology, we identified at least 40 
palm trees suspected of infestation.

In terms of palm tree detection from downloaded images, 
the palm tree aerial detector achieved a performance of 0.50 mAP and the street-level palm detector achieved mAP of 0.90.
The palm tree health classifier achieved an F1 score of 0.84, precision of 0.83, recall of 0.85, and AUC of 0.948.

To evaluate the potential of palm tree mapping from street view images we chose a small town highly populated with palms as a use case, in order to demonstrate the strength of the proposed method in identifying infested palms, specifically in urban environments. 
We download 3,209 panorama images from Google Street View of the small town of Omer, Israel.
We were highly successful in identifying palm trees in 2,609 out of the 3,209 panoramas (see Figure \ref{fig:omer}), even in this urban setting. 

\begin{figure}[h!]
\captionsetup[subfigure]{justification=centering}

    \centering
    \begin{subfigure}[t]{0.48\textwidth}
        \includegraphics[width=\textwidth]{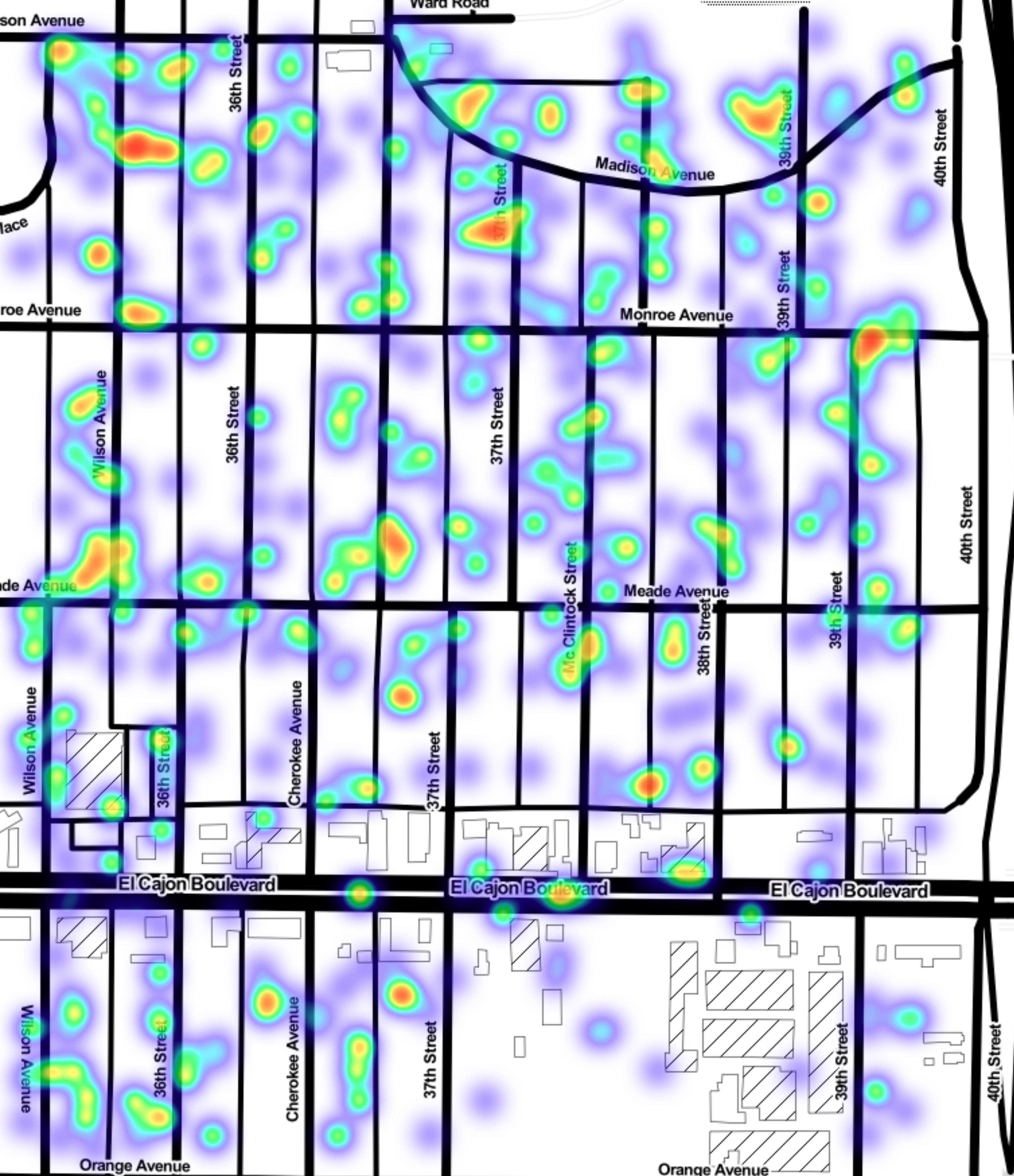}
        \caption{An aerial palm detection heat map.}
        \label{fig:naerial}
    \end{subfigure}
    ~ 
    \begin{subfigure}[t]{0.48\textwidth}
        \includegraphics[width=\textwidth]{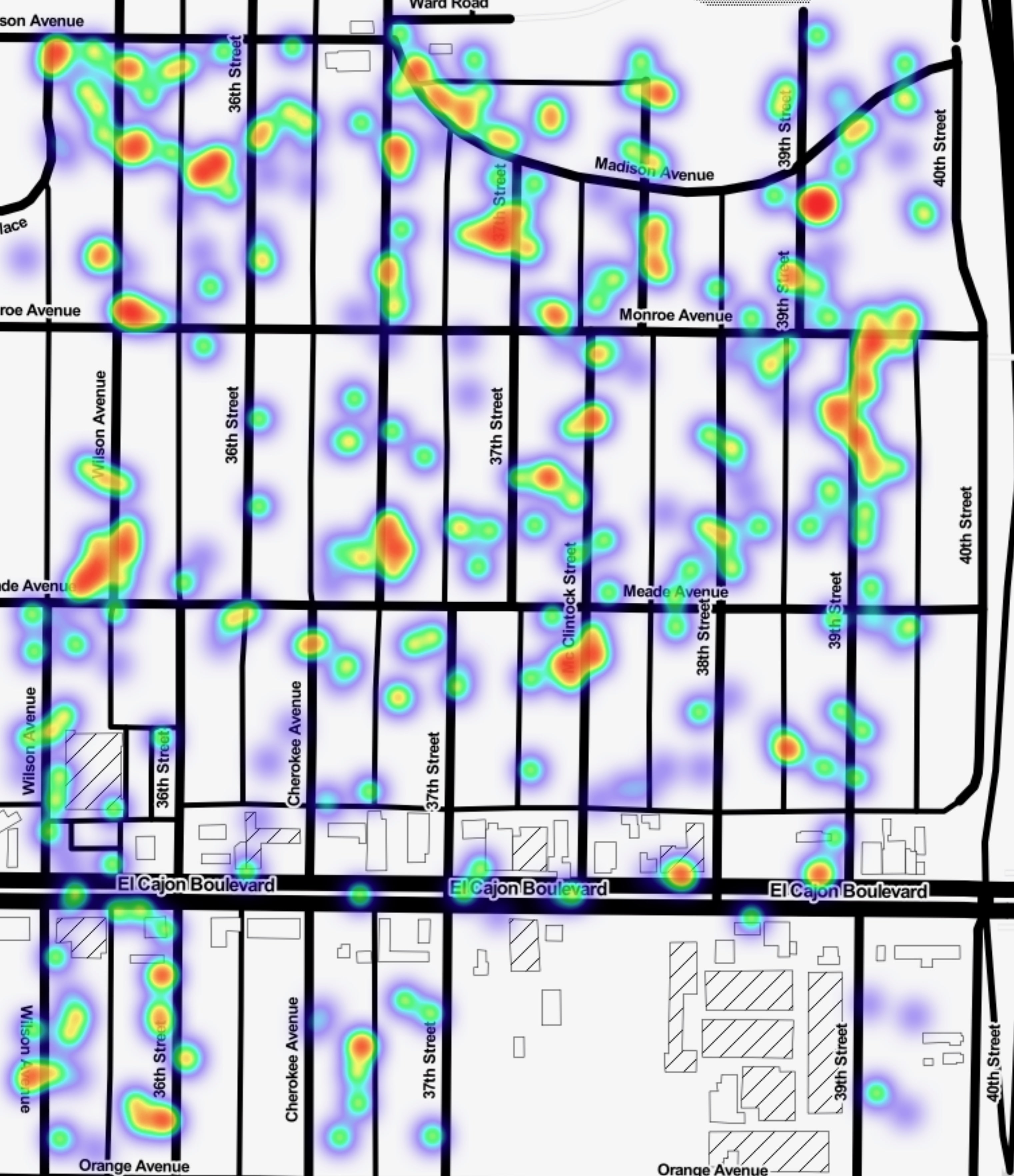}
        \caption{A street view palm detection heat map.}
        \label{fig:nstreet}
    \end{subfigure}
    \caption{Street vs aerial detection heat map presented using  Normal Heights Village, San Diego, CA}
    \label{fig:heat}
 \end{figure}

Next, we demonstrate that by using aerial images we reduce the number of required street view images needed for mapping palm trees in urban areas, significantly reducing our search space.
For example, to map palms in a sample neighborhood, namely the Normal Heights Village, San Diego we used 546 aerial images.
In these 546 aerial images, we detected 756 palm trees.
To map all Normal Heights Village areas using Street View only 1,136 panorama images are required, where each panorama is essentially composed of four images (the API only returns a 90-degree point of view each time).
Thus the search space is reduced in this example from 1136*4 street view images to 546 street view and 756 aerial images,
In Figure \ref{fig:heat} 
we show the comparison of palm tree detection using aerial images only (see Figure \ref{fig:naerial}) to palm tree detection using Street View after aerial detection (see Figure \ref{fig:nstreet})
It can be seen the palm tree detection results are highly similar in both cases (72\% of  palm trees detected using aerial images were confirmed as actual palm trees on street-level imagery), 
despite the pre-processing step of search space reduction by aerial imagery.

\begin{figure}[h!]
\captionsetup[subfigure]{justification=centering}

    \centering
    \begin{subfigure}[t]{0.3\textwidth}
        \includegraphics[width=\textwidth]{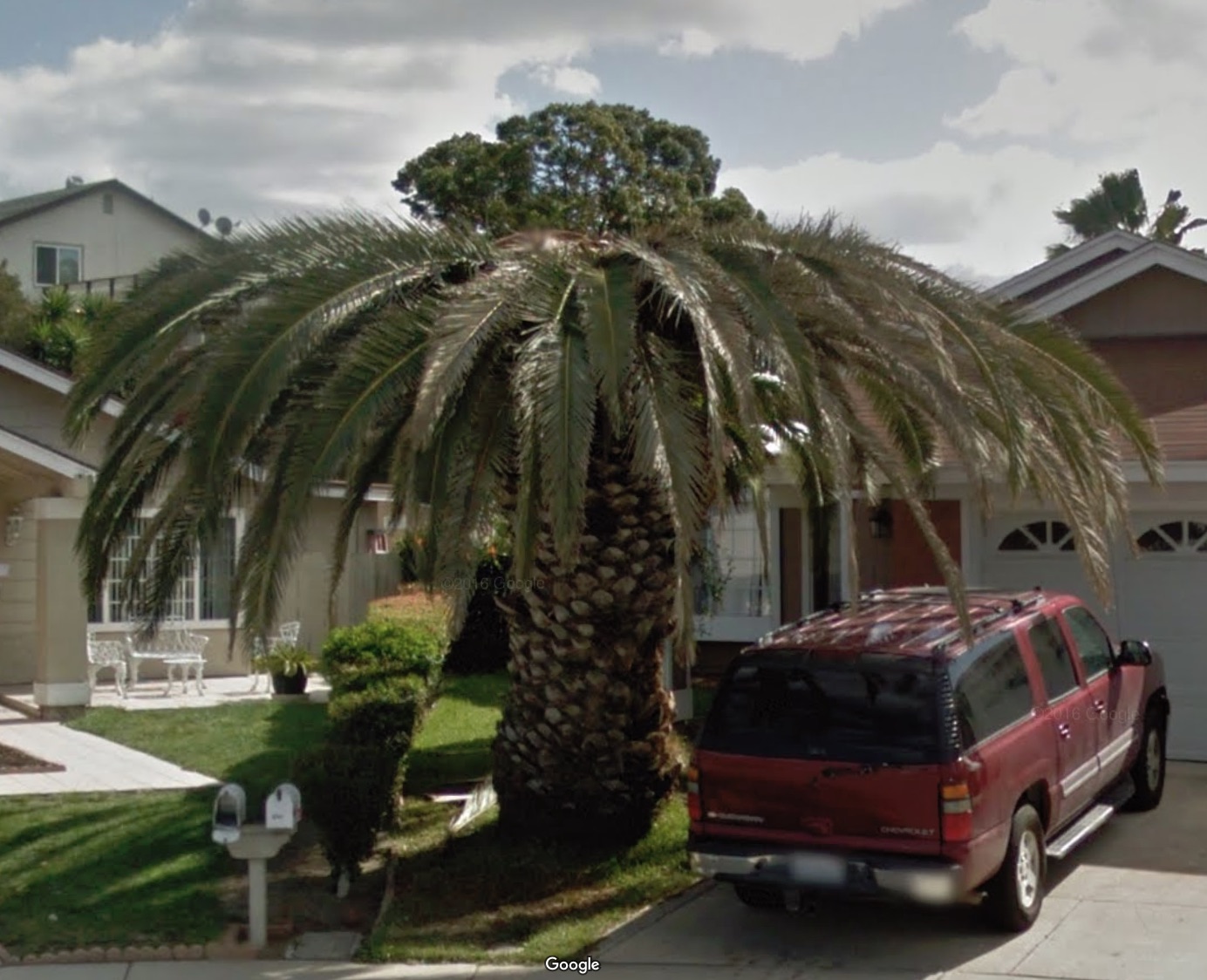}
        \caption{2281 Fantasy Lane, San Diego, California (February ,2016)}
        \label{fig:palm1}
    \end{subfigure}
    ~ 
    \begin{subfigure}[t]{0.3\textwidth}
        \includegraphics[width=\textwidth]{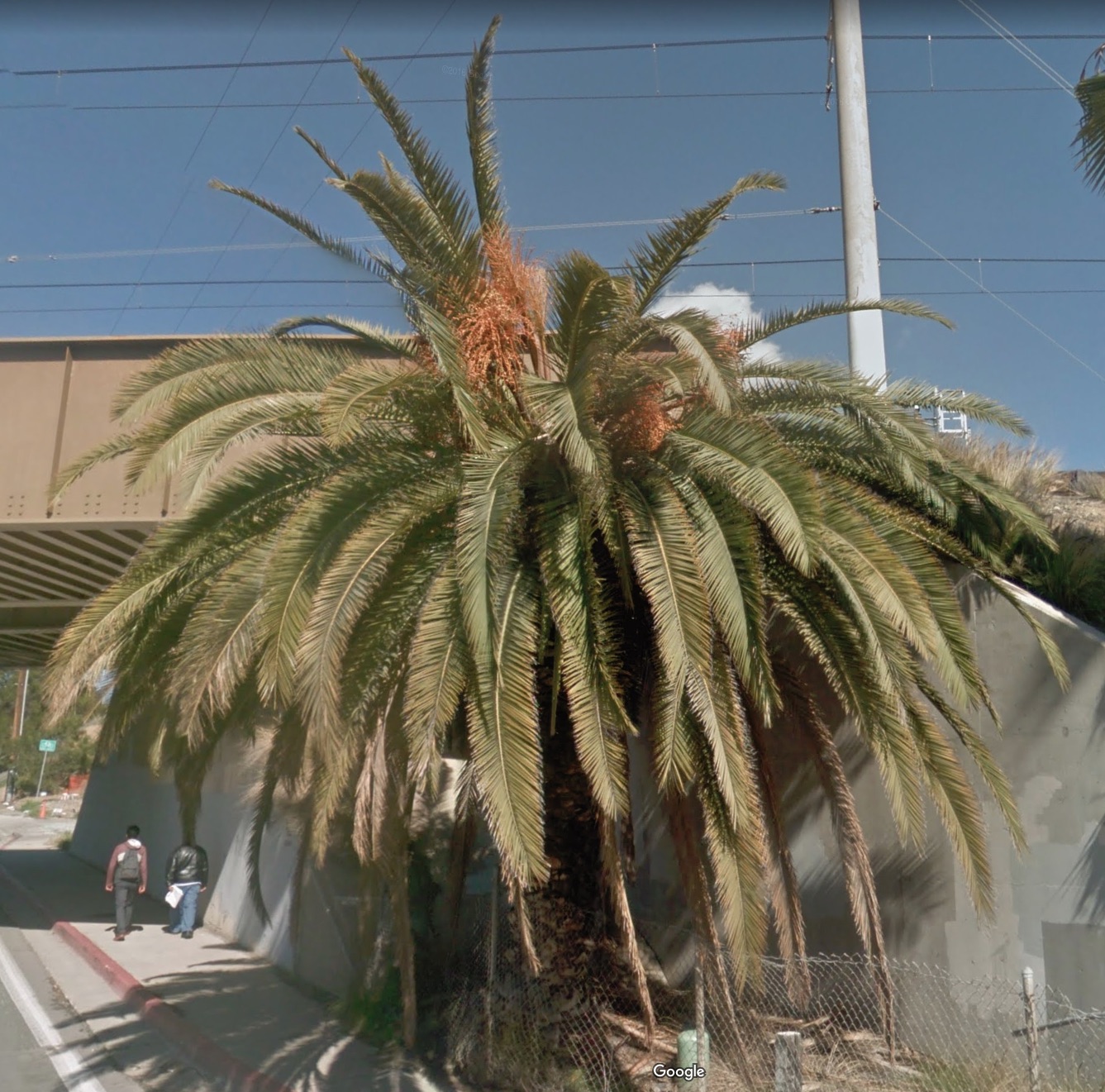}
        \caption{2464 E Beyer Blvd San Diego, California (February ,2016)}
        \label{fig:palm2}
    \end{subfigure}
    ~ 
    \begin{subfigure}[t]{0.3\textwidth}
        \includegraphics[width=\textwidth]{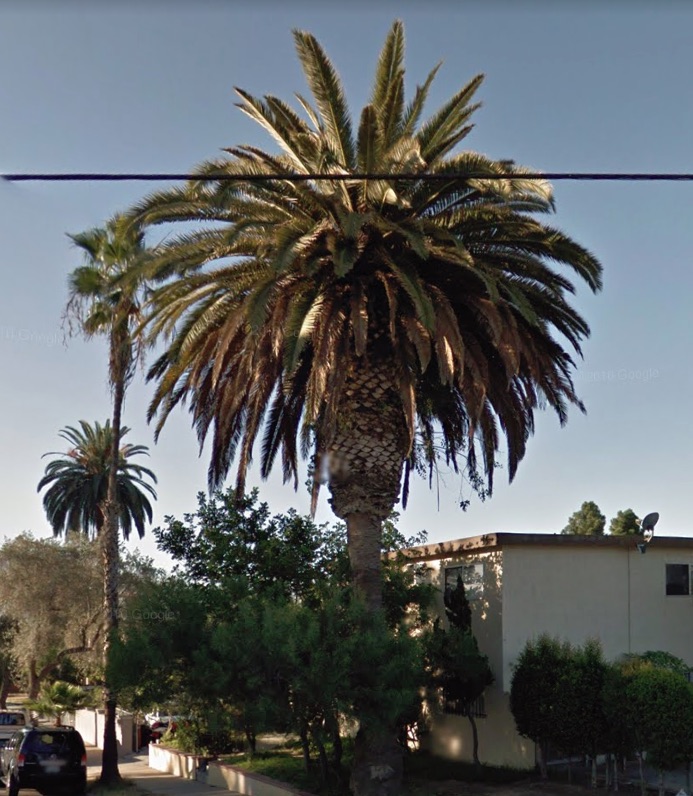}
        \caption{241 W. Park Ave., San Diego (November, 2015)}
        \label{fig:palm3}
    \end{subfigure}
    
\caption{Google street view images of the infested palm trees presented by \citet{hodel2016south}.}
        \label{fig:palms}

\end{figure}

As a proof of concept, we also demonstrate that we can use our proposed methodology to find actual infested trees in urban areas, using street view imagery.
Specifically, out of the four infested palm trees described by \citet{hodel2016south} we able to find the exact physical location of three of them (See Figure \ref{fig:palms}).
The location of the fourth tree was described with a general location description instead of an address, so could not be confirmed.
\citet{hodel2016south} reported infested palm trees are dated to March 2016.
The nearest images by date in Google street view were dated for two trees to February 2016 and the third three were dated to November 2015.
Additionally, we found all three palm trees that had specified location in an online report \citep{SouthAme35:online} (see Figure \ref{fig:palms2}).
Our classifier classified five of the six trees as infested.
Additionally, out of 5,008 detected palms, in the same area, the classifier detected additional 13 infested palm trees, of which we identified eight at advanced infestation stages.

\begin{figure}[h!]
\captionsetup[subfigure]{justification=centering}

    \centering
    \begin{subfigure}[t]{0.3\textwidth}
        \includegraphics[width=\textwidth]{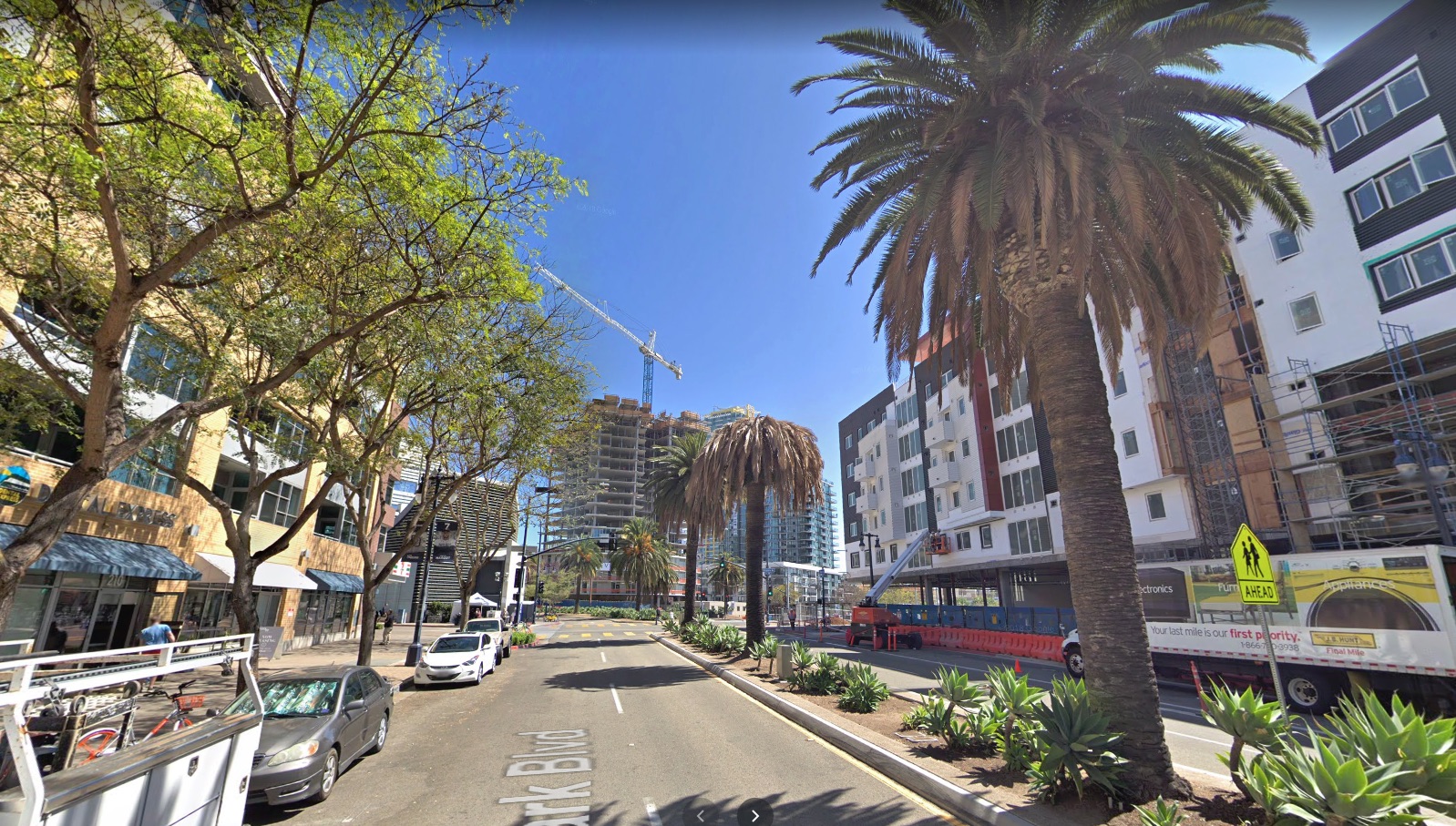}
        \caption{225 Park Blvd San Diego, California (April ,2018)}
        \label{fig:palm3}
    \end{subfigure}
    ~ 
    \begin{subfigure}[t]{0.3\textwidth}
        \includegraphics[width=\textwidth]{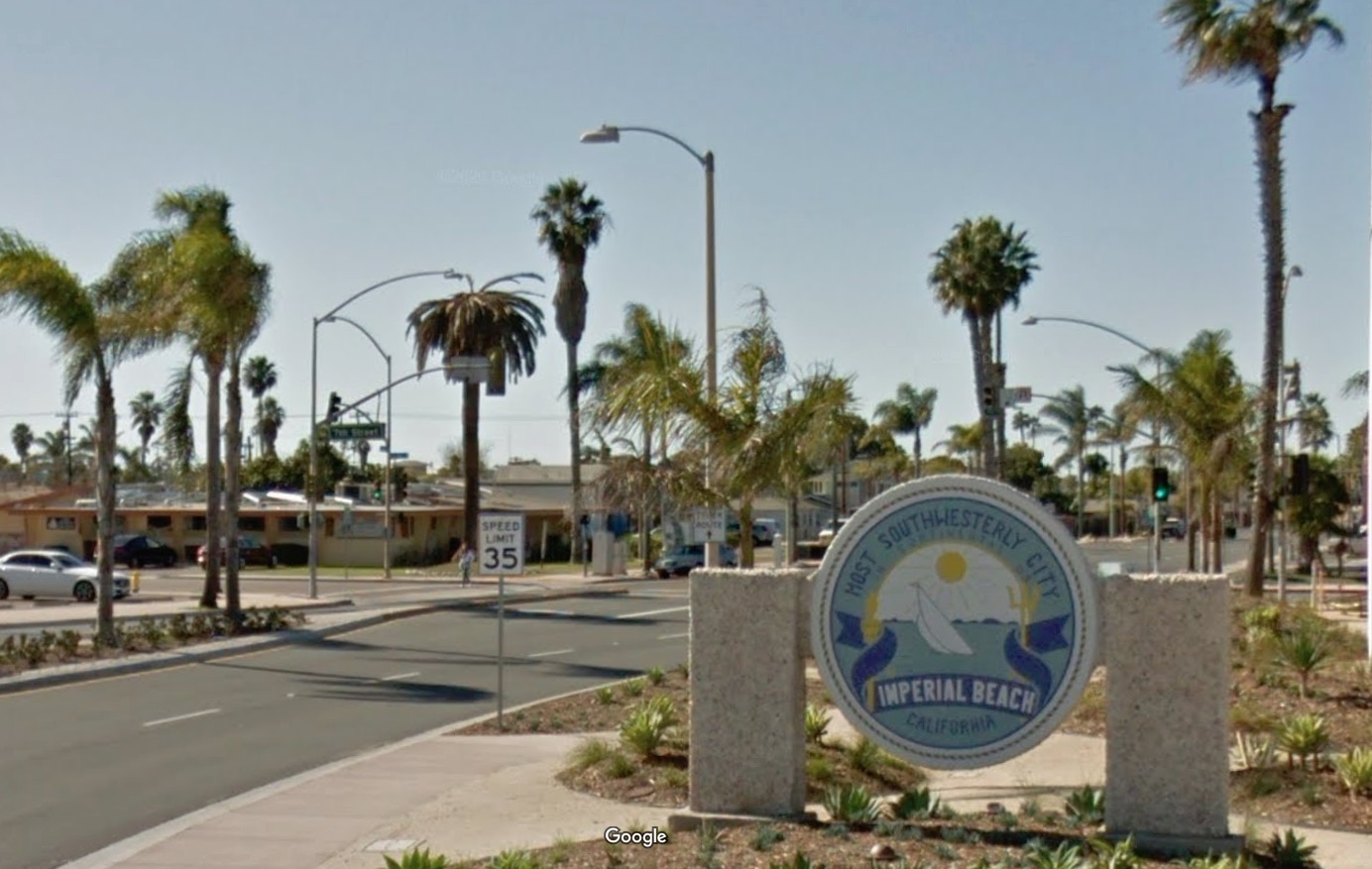}
        \caption{7th St Imperial Beach, California (January ,2018)}
        \label{fig:palm4}
    \end{subfigure}
    ~ 
    \begin{subfigure}[t]{0.3\textwidth}
        \includegraphics[width=\textwidth]{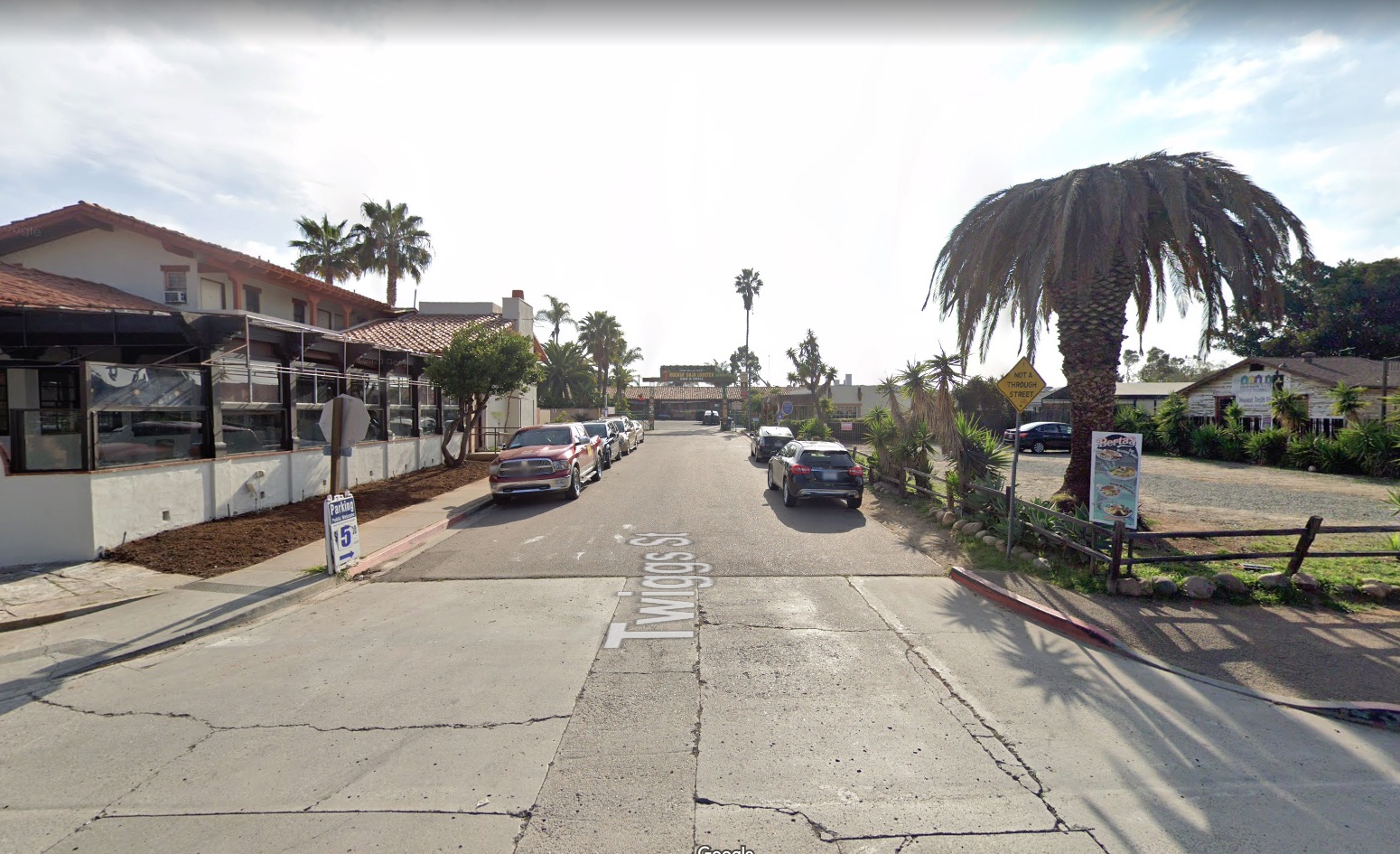}
        \caption{2572 Congress Street, San Diego, CA, USA (February, 2020)}
        \label{fig:palm5}
    \end{subfigure}
    
\caption{Google Street View images of the infested palm trees presented by Aguilar Plant Care \citep{SouthAme35:online}}
        \label{fig:palms2}

\end{figure}

\begin{figure}
\captionsetup[subfigure]{justification=centering}

    \centering
            \begin{subfigure}[t]{0.4\textwidth}
        \includegraphics[width=\textwidth]{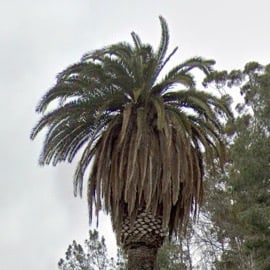}
        \caption{4452 Taylor St San Diego, California (May ,2019)}
        \label{fig:palm8}
    \end{subfigure}
     \begin{subfigure}[t]{0.4\textwidth}
        \includegraphics[width=\textwidth]{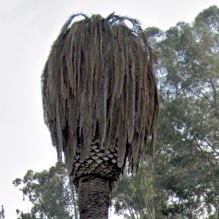}
        \caption{4452 Taylor St San Diego, California (February ,2020)}
        \label{fig:palm9}
    \end{subfigure}
    \begin{subfigure}[t]{0.4\textwidth}
        \includegraphics[width=\textwidth]{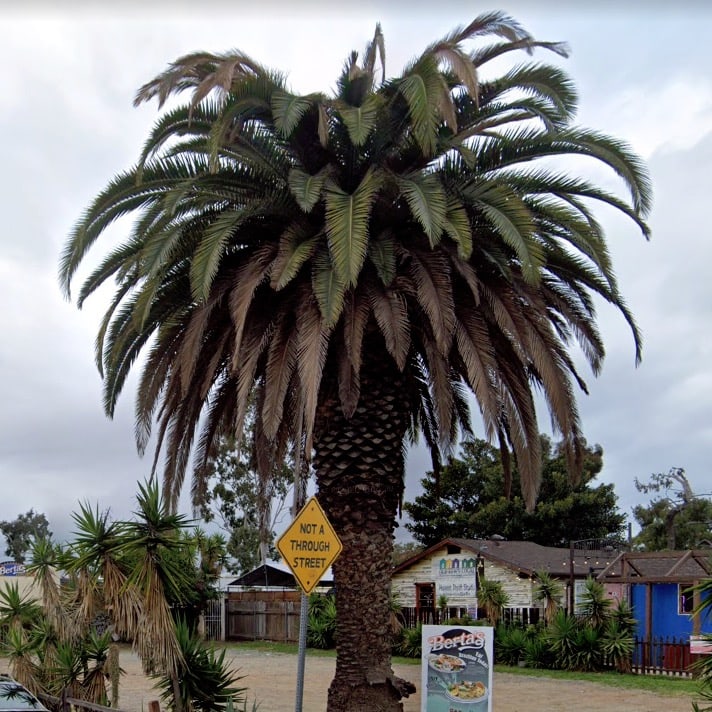}
        \caption{2572 Congress St San Diego, California (April ,2019)}
        \label{fig:palm4}
    \end{subfigure}
    \begin{subfigure}[t]{0.4\textwidth}
        \includegraphics[width=\textwidth]{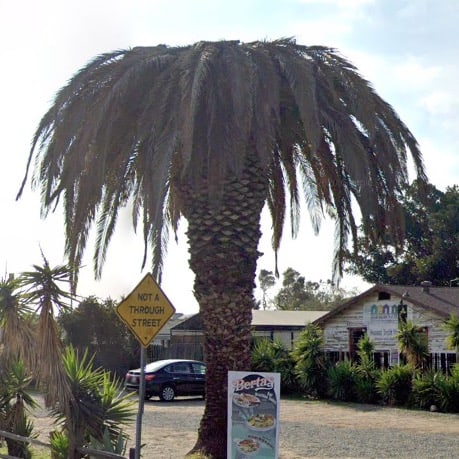}
        \caption{2572 Congress St San Diego, California (February ,2020)}
        \label{fig:palm5}
    \end{subfigure}
        \begin{subfigure}[t]{0.4\textwidth}
        \includegraphics[width=\textwidth]{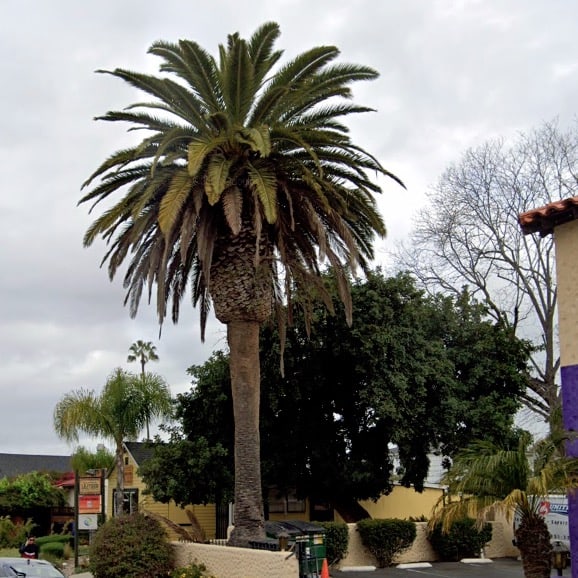}
        \caption{2505 Congress St San Diego, California (April ,2019)}
        \label{fig:palm6}
    \end{subfigure}
        \begin{subfigure}[t]{0.4\textwidth}
        \includegraphics[width=\textwidth]{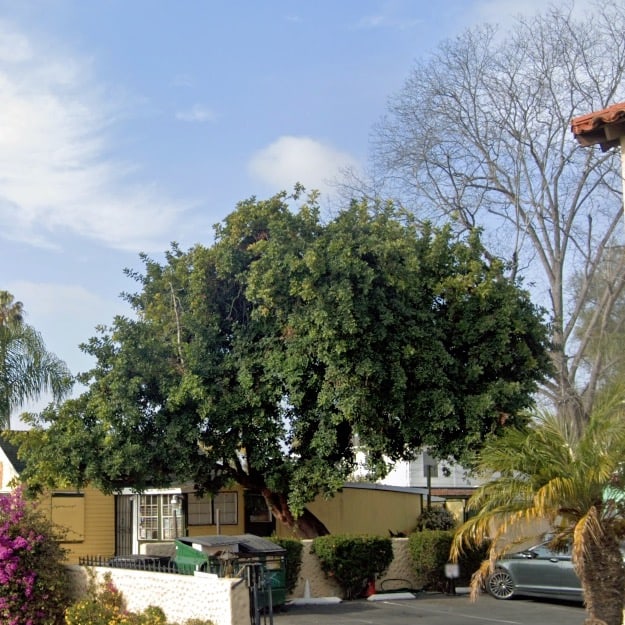}
        \caption{2505 Congress St San Diego, California (February ,2020)}
        \label{fig:palm7}
    \end{subfigure}

\caption{Examples of detected infested palm trees, in different stages of the infestation displayed by Google Street View Images. }
        \label{fig:palms2}

\end{figure}

\begin{figure}[h!]
\captionsetup[subfigure]{justification=centering}

    \centering
    \begin{subfigure}[t]{0.48\textwidth}
        \includegraphics[width=\textwidth]{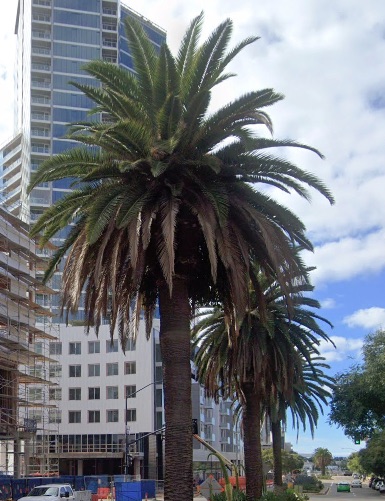}
        \caption{225 Park Blvd San Diego, California (November ,2017)}
        \label{fig:palm10}
    \end{subfigure}
    ~ 
    \begin{subfigure}[t]{0.48\textwidth}
        \includegraphics[width=\textwidth]{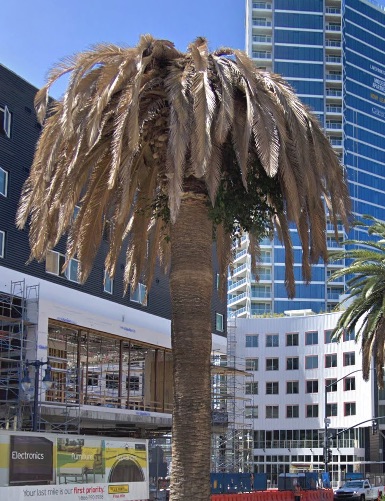}
        \caption{225 Park Blvd San Diego, California (April ,2018)}
        \label{fig:palm11}
    \end{subfigure}

\caption{Identified infested palm tree and the its latest images that identified as not infested. }
        \label{fig:before-after}

\end{figure}

We also wanted to demonstrate the potentially efficiently finding newly infested trees using aerial and street-level imagery combined.
We retrieved 22,438 aerial images that lead us to 54,781 street view images.
From these images, we found 36,001 palm trees from which 109 were classified as infested from which we identified 24 as an infestation at an advanced stage.
Additionally, we demonstrate that our method can be utilized to find  the time interval during which the palm tree was infested.
Observing Figure \ref{fig:before-after} we see that according to our classifier in November 2017 (see Figure \ref{fig:palm10}) the palm tree still was not infested.
However, in April 2018 (see Figure \ref{fig:palm11}) we identified infestation.

\section{Discussion}
\label{sec:dis}
We propose a novel framework for large scale mapping and detection of Red Palm Weevil infested palm trees, 
using state-of-the-art deep learning algorithms.
This large scale methodology may be of tremendous financial importance to countries around the world, help in massively saving agricultural fields, and assist in reducing risks of injuries in urban areas.
The proposed methodology relies on aerial and street view images available online.

First, we demonstrated how easily can Google Street View images can be used to map palm trees in small cities.
In a small town (Omer, Israel) chosen as a suitable use-case, we were able to detect thousands of palms.
We detected thousands of palm trees from Omer, Israel street view images. 
We found that in almost on every street in Omer there is a Palm tree, making this town extremely vulnerable to Red Palm Weevil infestation.
Nevertheless, we noted that not all streets of Omer are mapped by Google.
We propose that small towns, such as Omer, may choose to independent annual photoshoot the streets of their municipality, subsequently applying the offered methodology to 
the newest data to detect newly infested trees and perform preventive pestification to save trees.
Additionally, by performing more frequent mapping they can detect an active infestation and monitor its state.

Second, we show how we could reduce at least six-fold the number of street view images required for mapping in urban areas, by initially using rough aerial palm tree detection.
This translates into saving both money (using each street view image costs 0.007\$) and time in data collection and processing.
When mapping large areas using aerial images may save thousands of dollars.
Our results indicate that the majority (72\%) of aerial detected palm trees are of actual palm trees.
This number is likely even higher since some palms are not viewable from the street. 
Naturally, there are also false positives.
In other words, in most cases, there is no need to use all the area street images.
Full street mapping is only practical in small cities or when the municipality independently maps the streets, otherwise, the costs using street view could be very high\footnote{We estimate that fully mapping of San Diego area using only street view images should cost around 51,000\$ at the base pricing. The price probably would be lower, Google offers different pricing for large volumes but it is not specified in the documenting.}. 

Third, we illustrated that our method in general and palm tree health classifier, in particular, can successfully detect infested palm trees.
As a sample case, we detected two out three infected palm trees described by \citet{hodel2016south}.
The third reported tree showed only early stages of infestation on March 17, 2016, while the closest street view images dated November 2015. 
This may explain why it was not detected by the classifier as being infested, likely because it was not yet exhibiting visual signs of infestation.
The results indicated that deep learning can be used for detecting infested palm trees.
At the current status of the proposed model, it is able to detect severe and medium infections.
Future studies may focus on modifying the algorithms to detect various stages of infestation.
With more available data for training, as well as access to images of higher resolution, it may be possible to improve the accuracy of detection.
By using high-resolution trunk images it may possible to also detect early infestation signs in a variety of palm tree spices.

Fourth, we detected infested palm trees that were not presented in online reports, including areas that were not mentioned in any report.
Our results clearly show the potential of the method to monitor palm tree infestation worldwide utilizing minimal resources based only on online available data.
Urban mapping of palm trees may slow down the spread of infestation and flag areas for pestification. 
Moreover, theoretically- putting costs aside, it is possible to study the spread of the Red Palm Weevil over both time and space areas creating a map of their spreads. 
Such data may be used to predict the path of the spread in a future infestation.

\section{Research Limitations}
\label{sec:lim}
It is worth noting that the methods used in this study are prone to several limitations:
First, Google aerial images are not taken at the same time that of street view images, a fact that may lead to missing newly planted palm trees or detection of palm trees that were already cut down.
This of course can be solved by obtaining aerial images from multiple time points. 
However, currently, Google API does not support retrieving older aerial images nor the timestamp of the current images.
Second, occasionally the detected palm tree in aerial images does not have a line of sight from the street, thus we will not be able to acquire its street image. Nevertheless, we will at least know it is there, in contrast to using Street View images only.
Third, a single street view image may contain multiple palm trees, thus often it is hard to determine whether the same tree is detected by an aerial versus a street view palm tree detector.
Forth, street view images are not taken at constant time intervals, and only at sporadic time points.
This can lead to missing information in some areas at certain times.
Fifth, the number of images of infested palm trees available online is limited. 
Moreover, most of those images are of palms at advanced infestation stages.
This limits the performance of the infestation classifier both in terms of accuracy, as well as on different stages of infestation. 
Training the classifier on more data with infested trees on different stages should provide better results.
Sixth, there are many types of palm trees and the symptoms of an infestation can be different.
Currently, most of the photos we found of infested palm trees are of Cannery Palm trees.
With additional data, a specific classifier can be created for each type of palm tree.

\section{Conclusions}
\label{sec:con}
Red Palm Weevil has spread across the world damaging and destroying countless palm trees and date crops.
In this study, we developed a novel automatic framework for  detection and monitoring of Red Palm Weevil infestation by analyzing over 47,000 aerial and 61,000 street views. 
We first filter the enormous search space by detecting palm trees in aerial images; then we further analyze those aerial detected palms using street level images.
We demonstrated that this information can be utilized in order to map palm trees in urban areas and to detect and monitor Red Palm Weevil infestation.

We demonstrated that using deep learning algorithms and online available data provides an automatic and cost-effective solution for monitoring and detection of pest infestations.
Such a solution can help ministries of agriculture and governments worldwide to fight and contain the spread of Red Palm Weevil in a cost-effective fashion.
Our method can be instantly deployed to any scale of need and every geographic location where street view level imagery is accessible.
Standard commonly existing procedures for monitoring infested palms require specialized equipment, time, cost and effort, thus not very effective for large scales.
In the current state of the Red Palm Weevil spread such a tool is essential for confronting against the Red Palm Weevil damage.

\section{Data availability}
The data contains mostly images from Google Maps and Street View, the data cannot be distributed according to Google terms of service.
However, the images can be accessed using Google API.

\section{Acknowledgements}
We thank Valfredo Macedo Veiga Junior (Valf) for designing the infographic illustration and Omer Tsur - AirWorks aerial photography for providing the Deganya Alef aerial photo.

\bibliographystyle{unsrtnat}
\bibliography{sample-bibliography}

\end{document}